\documentclass[preprint,12pt,authoryear]{elsarticle}

\usepackage[utf8]{inputenc}
\usepackage{url}

\usepackage{amssymb}
\usepackage{adjustbox} 
\usepackage{multirow} 

\usepackage[noabbrev]{cleveref}

\usepackage{threeparttable} 
\usepackage{bm} 

\usepackage{eurosym} 

\usepackage{microtype}

\usepackage{xcolor}

\usepackage{float}
	
\journal{arXiv}

\begin{document}

\begin{frontmatter}

\title{Human vs. supervised machine learning: \\ Who learns patterns faster?}

\author{Niklas Kühl}
\author{Marc Goutier}
\author{Lucas Baier}
\author{Clemens Wolff}
\author{Dominik Martin}

\address{Karlsruhe Institute of Technology (KIT)\\ Kaiserstr. 89, 76133 Karlsruhe, Germany\\ \{niklas.kuehl, marc.goutier, lucas.baier, clemens.wolff, dominik.martin\}@kit.edu}

\begin{abstract}

The capabilities of supervised machine learning (SML), especially compared to human abilities, are being discussed in scientific research and in the usage of SML. This study provides an answer to how learning performance differs between humans and machines when there is limited training data. We have designed an experiment in which 44 humans and three different machine learning algorithms identify patterns in labeled training data and have to label instances according to the patterns they find. The results show a high dependency between performance and the underlying patterns of the task. Whereas humans perform relatively similarly across all patterns, machines show large performance differences for the various patterns in our experiment. After seeing 20 instances in the experiment, human performance does not improve anymore, which we relate to theories of cognitive overload. Machines learn slower but can reach the same level or may even outperform humans in 2 of the 4 of used patterns. However, machines need more instances compared to humans for the same results. The performance of machines is comparably lower for the other 2 patterns due to the difficulty of combining input features.

\end{abstract}

\begin{keyword}	
supervised machine learning; human learning; cognitive psychology; pattern recognition; small sample size; experimental study
\end{keyword}

\end{frontmatter}

\section{Introduction}
\label{sec:introduction}
Supervised machine learning (SML), with its capabilities to support---or even replace---human workers in their daily tasks, is omnipresent in current discussions. While research is investigating the capabilities of SML in a broad range of areas, for example image classification \citep{he2016deep} or speech recognition \citep{hinton2012deep}, tasks where machine learning models outperform humans are increasing \citep{grace2018will}. For instance, in the field of autonomous driving, replacing humans as drivers is supposed to happen sooner or later, although autonomous vehicles are not (yet) able to completely substitute humans in this task \citep{casner2016challenges}. In the long run, several studies foresee humans being completely substituted by machines in many tasks, including their work processes \citep{muller2016future,makridakis2017forthcoming}. But can this development be observed across all tasks? Current examples of (supervised) machine learning models outperforming humans are mainly present in areas where a high amount of training data is available, for example billions of played ``Go'' games \citep{chang2016google} or millions of labeled images \citep{russakovsky2015imagenet}. 

In real life, however, often only limited ``training'' data is available \citep{baierchallenges}---sometimes just a single instance \citep{lee2015neural}. In this article, we are especially interested in learning patterns by humans and machines in data with only a few instances on a given task. While there is theoretical work in the field of \textit{inductive programming} \citep{muggleton1991inductive, olsson1995inductive}, which aims to design techniques capable of capturing patterns with few examples, empirical work in the comparison to human learning is still rare. The research stream of Cognitive Sciences has been investigating learning processes of humans (and recently also machines) \citep{favela2017cognition}, especially with a focus on understanding and mimicking the human brain \citep{thagard2005mind,dupoux2018cognitive}. As a subfield, computational cognitive science has been studying the similarities and differences between human and machine learning in the past two decades \citep{kogler2017celebration, tenenbaum2011grow, lake2015human, griffiths2007google, griffiths2010probabilistic, lake2017building, lucas2015rational}.

However, the investigation of the learning curves, meaning the relation of required training samples and the resulting performance \citep{perlich2003tree} of humans in comparison to SML models, is a topic that has not yet been investigated. This investigation is of major importance, as it needs to be considered whether humans or machines will be performing a task, especially in future, SML-based applications. It gives more insights to the question about which entity learns more efficiently \citep{hernandez2017measure}. For instance, in the field of healthcare, data labeling by physicians is extremely costly. From an economic perspective, it might be questionable to use supervised machine learning models in healthcare because the labeling cost can exceed the machine's saving potential \citep{raghupathi2014big}. To give first insights into comparing the learning curves of humans and machines with limited training data, we phrase our general research question (GRQ) as follows:

\begin{center}
	\textit{GRQ: How does the learning performance of humans and supervised machine learning models differ with limited training data?}
\end{center}

In academia, direct comparisons between humans and supervised machine learning models performing the same task are still rare \citep{hernandez2017evaluation}. Besides the aspect of limited training data, it must be of fundamental interest for researchers to gain a better understanding of which tasks can possibly be undertaken by a supervised machine learning model, as well as the precise conditions that apply \citep{witten1994comparing, adler2007comparing, marcus2018algebraic}. As this work reinforces, there are infinite possibilities of tasks and task characteristics. To provide a starting point for research endeavors in the supervised field, we explore one special scenario. The chosen task for humans as well as machines is identifying patterns with limited training samples (5 to 50 instances). To measure the human performance on this task, we conduct a lab experiment with 44 participants where four different patterns need to be identified. We then apply different machine learning algorithms on the same patterns and subsequently evaluate and compare the results. 

The remainder of this work is structured as follows: In the next Section, we present the necessary fundamentals and related work in the fields of human and machine learning (\Cref{sec:relatedwork}). Next, we define the overall task characteristics (\Cref{subsec:prerequisites}) and elaborate on our methodological focus for the experiment design (\Cref{subsec:experimentdesign}). In \Cref{sec:results}, we report the isolated results (humans and machines respectively) of the task performance and subsequently introduce the comparison. Finally, we discuss implications (\Cref{sec:discussion}) and conclude the study (\Cref{sec:conclusion}). 
\section{Fundamentals and related work}
\label{sec:relatedwork}

This article assesses and analyzes learning performances of humans and machines. To sketch the foundation for this endeavor, we first give an overview of current research in learning, segmented into learning of humans (\Cref{subsec:humanlearning}), machines (\Cref{subsec:machinelearning}), and research on their comparison (\Cref{subsec:humanmachinelearning}). 

\subsection{Human learning}
\label{subsec:humanlearning}

Scientific research on human learning started in the second half of the 19$^{th}$ century. In one of the first books about learning, \cite{ebbinghaus1885gedachtnis} postulated the concept of a learning curve, as the subject group's learning progress flattened over time. \cite{kotovsky1973empirical} started analyzing humans learning patterns on a large scale. Human learning is currently divided into three main learning theories: Cognitive psychology, social cognitive theory, and sociocultural theory \citep{ormrod2004human}.

For this article's research topic---analyzing human learning with small sample sizes and comparing it to SML---the most relevant research field is cognitive psychology. We leverage phenomena from this area to provide possible explanations for human learning patterns. Cognitive psychology is ``the study of how people perceive, learn, remember, and think about information'' \citep[p. 3]{sternberg2016cognitive}. The research on cognitive psychology includes studying mental phenomena, such as visual perception, object recognition, attention, memorization, knowledge, speech perception, judgment, and reasoning. To explain such phenomena, cognitive psychology has recourse to neuroscience and its knowledge of brain functioning \citep{eysenck2015cognitive}.

In turn, social cognitive theory \citep{rosenthal1978social} includes many ideas from cognitive psychology, but focuses on how humans learn from other human beings through watching and imitating their behavior. The theory suggests that humans can control their own learning. This differs from behaviorism, a now dated theory which led to social cognitive theory and in which learning is solely the result of stimulus–response relationships \citep{ormrod2004human}. Learning from others also has the benefit of learning quicker by making fewer mistakes compared to learning from own experiences \citep{bandura1986social}.

Sociocultural theory stresses the importance of society and culture in learning. Learning a sociocultural tool like a language is not only useful for communication, but also supports humans in their thinking development \citep{vygotsky1964thought}. In contrast with social cognitive theory, humans do not only learn from each other but also work together towards goals that cannot be achieved by individuals. The research focuses on the interaction of children and parents. Children's individual development of capabilities are usually related to interactions with their parents. Additionally, caregivers like parents can broaden a child's problem-solving abilities and stimulate cognitive growth by assisting them to solve more difficult tasks that they would otherwise not be able to accomplish \citep{vygotsky1980mind}.

A few number of studies studied how humans learn from a few instances and have also built computational models to understand human few-shot learning \citep{vul2014one, lieder2012burn}. However, their aim was to re-engineer the human learning process, while we aim to show empirically how the two entities perform in a direct comparison on limited training data.

\subsection{Machine learning}
\label{subsec:machinelearning}

The capabilities of machines have been discussed from various perspectives, including their abilities to capture knowledge \citep{lieto2018knowledge}, think \citep{hoffmann2010can}, feel \citep{velik2010machines, o2012build, osuna2020development}, be creative \citep{veale2010computational} and making morally good decisions \citep{tavani2011can, yilmaz2017computational}. The process of how machines obtain their knowledge in the first place is addressed in the area of machine learning. Machine learning describes a set of techniques commonly used to solve a variety of real-world tasks with the help of computer systems that can learn to solve a task instead of being explicitly programmed to do so \citep{Koza1996}. In general, we differentiate between unsupervised, reinforcement, and supervised machine learning \citep{Jordan2015}.

Unsupervised machine learning comprises methods and algorithms that reveal previously unknown data patterns. Consequently, unsupervised learning tasks do not necessarily have a ``correct'' solution, because there is no ground truth \citep{wang2009cvap}. In the area of reinforcement learning, rewards and punishments allow the model to learn continuously over time with many learning instances. The focus is on a trade-off between an uncharted environment’s exploration and the knowledge base’s exploitation \citep{kaelbling1996reinforcement}.

In this study, we mainly focus on supervised machine learning, because the most widely used methods are supervised \citep{Jordan2015}. It therefore seems to be a promising starting point. In respect of supervised machine learning, learning means that a series of examples (``past experience'') is used to build knowledge about a given task \citep{Dietterich1996}. Although statistical methods are used during the learning process, manual adjustment and rule or strategy programming to solve a task are not required. In more detail, (supervised) machine learning techniques aim to build a model by applying an algorithm to a set of known data points to gain insight into an unknown set of data \citep{hastie2017}. Typically, supervised machine learning models rely on large amounts of data to work properly. First techniques, not all directly related to SML, aim to reduce the required amount with different techniques, namely inductive programming \cite{olsson1995inductive, schmid2011inductive}, genetic programming \citep{banzhaf1998genetic}, active learning \cite{settles2009active}, semi-supervised learning \cite{lin2010semi}, combinations of both \citep{rhee2017active}, external memories \cite{vinyals2016matching} or one-trial learning \citep{feng2019simulating}. 

In terms of a supervised machine learning model’s ``creation'' procedures, the proposed processes vary slightly in their definition of the phases, but generally employ the three main phases: model initiation, performance estimation, and deployment \citep{hirt2017end}. During the model initiation phase, a task is defined, the data is prepared and processed, and a suitable machine learning algorithm is chosen. During performance estimation, various parameter permutations describing the algorithm are validated and a suitable configuration is selected based on its performance when solving a specific task. Lastly, the model is deployed and put into practice to solve a task related to previously unseen data. 

\subsection{Human vs. machine learning}
\label{subsec:humanmachinelearning}

When it comes to the comparison of human and SML, \cite{hernandez2017measure} motivates the comparison of natural and artificial intelligence in the first place. The field of Neuroscience \citep{floraz2015michael, rajalingham2018large, hutto2015looking} aims to understand the learning of humans and the facilitation with machines on a theoretical level. The precise capturing of the related learning curves have been analyzed theoretically and empirically for humans and machine learning techniques separately in different domains, e.g. creativity tests \citep{oltecteanu2016artificial}, face recognition \citep{adler2007comparing}, music prediction \citep{witten1994comparing} or cognitive research \citep{marcus2018algebraic}. In the field of computer vision, multiple comparisons of human and machines have been made \citep{elsayed2018adversarial, zhou2019humans, eckstein2017humans, peterson2018evaluating}. 

Apart from these specific domains and closer related to our study is the idea to build computer models capable of solving IQ tests \citep{hernandez2016computer}. While not using supervised machine learning, \cite{insa2011comparing} aim to compare reinforcement and human learning, however, they only regard small sample size of observations.

Human learning can be compared to machine learning based on various aspects. \cite{dubey2018investigating} focus on human priors for playing video games. In their experiment, they use an unknown video game that a human solves quite easily by using its priors on semantics, gravity, and objects. By reversing semantics and masking affordances, the human performance decreases drastically. The machine performance, represented by reinforcement learning algorithms, performs significantly better under the same conditions. Humans' prior knowledge is important when it comes to solving new problems quickly.

\cite{kim2019neural} does research on psychophysical phenomena---which can be found in human learning---in trained machine learning models. Gestalt phenomena are a part of human visual perception in which humans realize that the whole differs from the sum of its parts \citep{kohler1967gestalt}. They show that some neural networks are able to show one type of Gestalt phenomena under the proper circumstances. 

In hybrid intelligence \citep{dellermann2019hybrid}, humans' complementary strengths, like flexibility and common sense, are combined with those of machines, for example consistency and speed. This sociotechnological ensemble can overcome the current limitations humans and machines have. Another way to combine human and machine abilities is to treat machines as teammates \citep{burr2018analysis, smart2018human, seeber2019machines}. This could increase work speed and lead to better decision-making by detecting negative cognitive biases.

In conclusion, a direct comparison of human learning and supervised machine learning for the same task with limited training data availability still remains a research gap and is addressed in this work.

\section{Methodology}
\label{sec:methodology}

With the related work at hand, we outline the methodology used in this article. We first set necessary prerequisites (\Cref{subsec:prerequisites}) and then elaborate on the experiment's design (\Cref{subsec:experimentdesign}).

\subsection{Prerequisites}
\label{subsec:prerequisites}

Before discussing the experiment's design of comparing human and machine learning, we need to set the prerequisites for the task being solved in this experiment. As we require a controllable task with precise benchmarks for performance evaluation, a suitable candidate is supervised machine learning, which we utilize for this article. When it comes to choosing a meaningful task in the area of SML, there are many possible characteristics to describe it. To deduce the possibilities and reason our selection, we look at corresponding task characteristics (\Cref{subsubsec:task}) and subsequently outline our implementation of the chosen task (\Cref{subsubsec:taskimplement}).

\subsubsection{Task characteristics}
\label{subsubsec:task}

A learning curve depicts task performance based on experience. In our case, experience is measured by the amount of training data, more precisely by the number of training instances. Task performance is influenced by two main factors: the characteristics of the entity performing the task (humans or machines) and those of the task itself. Depicting a general learning curve for every type of task characteristic exceeds the scope of this article, and we have to limit the scope to an interesting selection of all possible tasks. For our supervised machine learning task, four task characteristics are of importance: input, output, instances, and features.

\paragraph{Input} The input describes the data the task is based on. It can differ by data type (e.g., numeric or binary) and by data representation (e.g., table, picture, or audio).
\paragraph{Output} A task also differs in the demanded output. Two types of output are relevant in this case: classification and regression. A classification determines whether each instance belongs to one of the predetermined classes, whereas the result of a regression is a continuous number.
\paragraph{Instances} The number of instances that are available for the learning process. 
\paragraph{Features} The instances of a task are described by a fixed number of distinct features.
\newline

To start the research endeavor, we select a task with a binary input, a binary classification as output, a small set of training instances, and a limited number of features. An overview of all task characteristics of interest and their implementation in this work can be found in \Cref{tab:characteristics}. To conclude, we update the general research question to our research question (RQ):

\begin{center}
	\textit{RQ: How and when do learning curves differ between humans and supervised machine learning models for small sample sizes, using a binary classification with limited binary features?}
\end{center}

Additionally, we define the following requirements for our task: it should not require prior knowledge and should use a balanced data set (same number of true and false instances) and should be solvable in a reasonable timeframe. The task should be represented in a suitable way for humans and machines, and it should be possible to depict the results in a learning curve.

\begin{table}[h]
	\begin{tabular}{|lp{2.2cm}|p{4.4cm}|p{4.4cm}|}
		\hline
		\multicolumn{2}{|l|}{Task characteristic} & Attributes                         & This work                          \\ \hline
		\multicolumn{1}{|c|}{}        & Data type                 & e.g. numeric data, binary data & binary data                       \\ \cline{2-4}
		\multicolumn{1}{|c|}{Input} & Data representation       & e.g. table, picture, audio         & picture (humans), table (machines) \\ \hline
		\multicolumn{2}{|c|}{Output}              & classification, regression         & binary classification              \\ \hline
		\multicolumn{2}{|c|}{Instances}           & number of instances                 & 5 to 50    \\ \hline
		\multicolumn{2}{|c|}{Features}            & number of features                 & 9 features   \\ \hline                     
	\end{tabular}
	\caption{Overview of the task characteristics of interest and their implementation in this work}
	\label{tab:characteristics} 
\end{table}

\subsubsection{Implementation of task characteristics}
\label{subsubsec:taskimplement}

As a last prerequisite, we have to agree on a task that satisfies our set of task characteristics and also complies with the additional requirements defined in \Cref{subsubsec:task}.

We use two suggestions in the field of intelligence tests as a foundation for our task, namely minimum intelligent signal tests (MISTs) and Raven's progressive matrices (RPMs): MISTs are binary questions that are used to quantify humanness \citep{mckinstry1997minimum, lupkowski2019minimum}. Compared to other intelligence tests, these questions do not require a complex answer, but only a simple yes or no, which satisfies our limitation on a binary output. However, the input is natural speech and not a set of a few, binary features. RPM \citep{raven2000raven} is a test of visual geometric objects, designed by a rule. The task is to complete the set of visual geometric objects by selecting an object out of six or eight options. Only one of the selectable objects matches the rule. RPMs have a graphical representation that can be reduced to a set of instances with a few binary features to get standardized instances. However, they lack a binary output. 

\begin{figure}[]
	\centering
	\includegraphics[width=\textwidth]{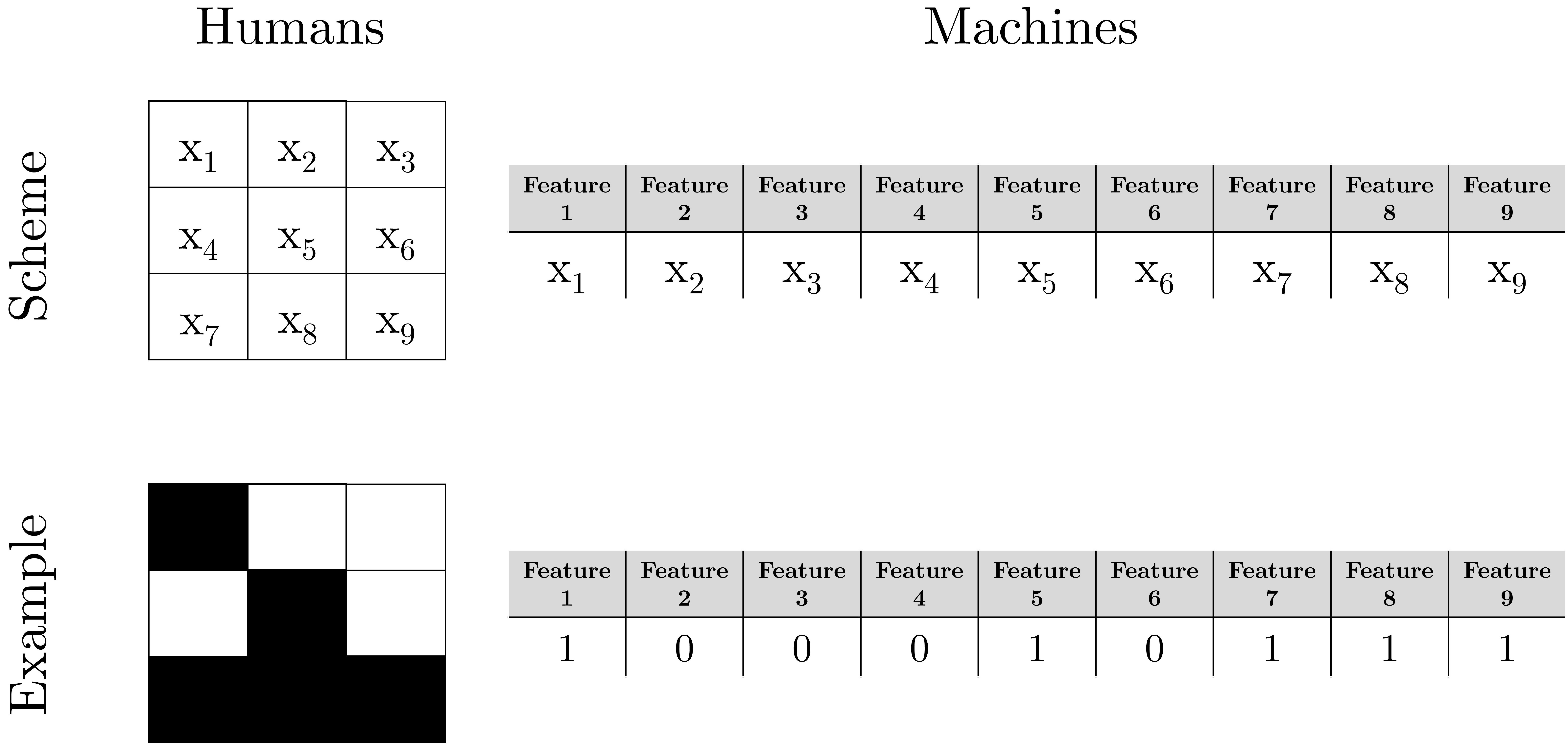}
	\caption{Schematic representation for humans and machines of an instance with features $x_1$ to $x_9$}
	\label{fig:schematic} 
\end{figure}

By combining these two tests, we define the following task: To have the same number of features, we use only 3x3 matrices with 9 elements (= 9 features). Every feature is binary. Accordingly, we have a set of $2^9$ = 512 different matrices. These matrices can be displayed as a picture with elements of black and white (for humans) or as a list of numbers with features of 1 and 0 (for machines). \Cref{fig:schematic} shows an example of how the same instance is represented for humans and machines respectively. Based on a rule regarding the feature value, we can classify the matrices: Some instances (matrices) fulfill the rule, therefore they are labeled as true, whereas all the other instances do not fulfill the rule and are labeled as false. We define four basic patterns as the four rules that define our classification task.

\paragraph{Diagonal} Matrices that fulfill the diagonal rule have at least one diagonal line that is labeled black, either starting in the upper left block and continuing to the lower right block, or stating in the lower left block and ending in the upper right block.	
\paragraph{Horizontal} Matrices that fulfill the horizontal rule have at least one horizontal row of only black elements.
\paragraph{Numbers} The numbers rule is satisfied if five elements in total are labeled black.
\paragraph{Symmetry} Symmetry describes axis symmetry, either to the middle column or the middle row of the matrix.		

\begin{figure}[]
	\centering
	\includegraphics[width=\textwidth]{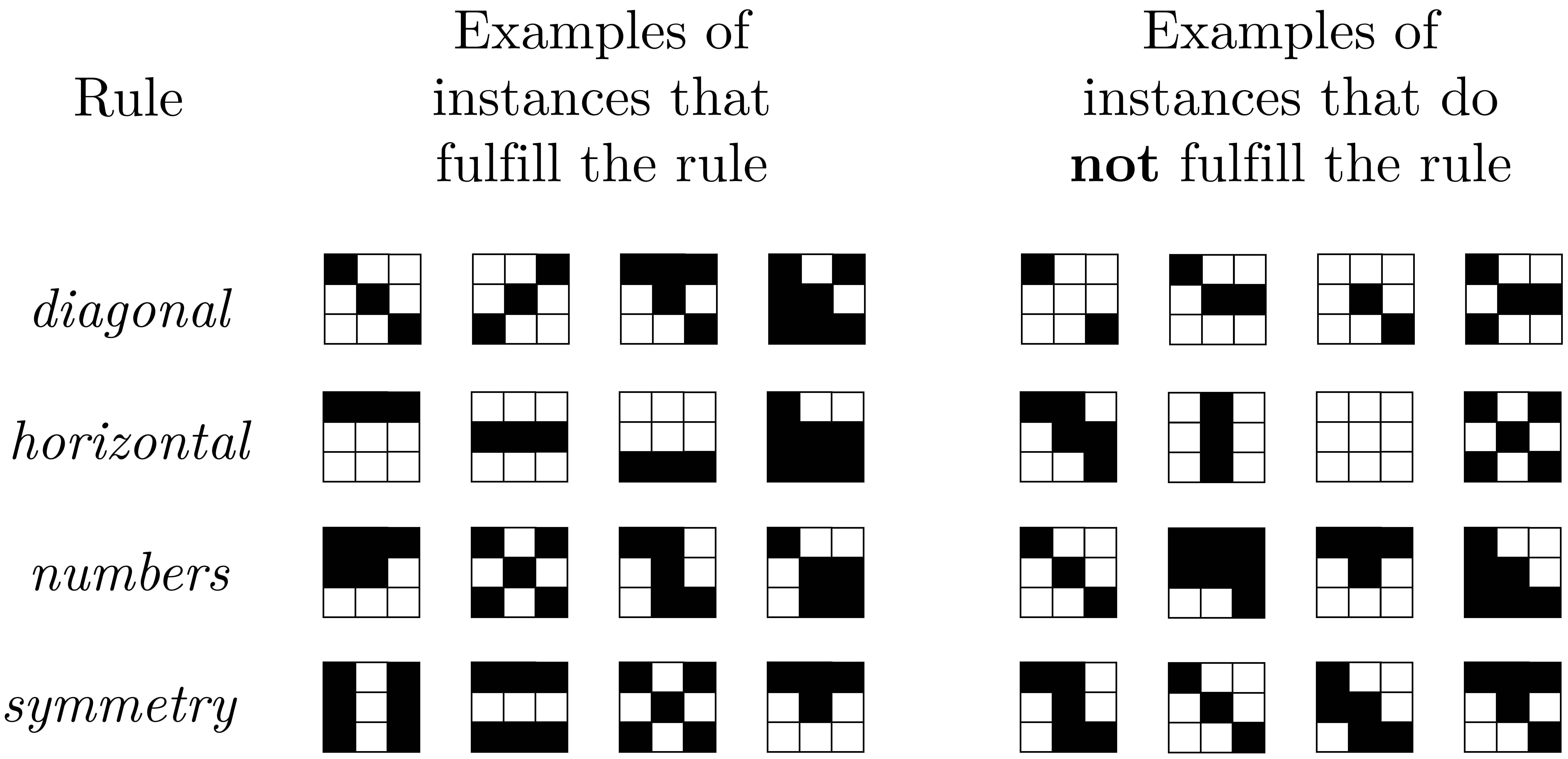}
	\caption{Exemplary instances for each rule}
	\label{fig:rules} 
\end{figure}

\subsection{Experiment design}
\label{subsec:experimentdesign}

The previously described task has to be adjusted to be conducted by humans (\Cref{subsubsec:humanexperiment}) as well as machines (\Cref{subsubsec:machineexperiment}) in an experimental setting. In the end, the results should render a learning curve. To generate a learning curve for a specific rule, we define a game with multiple rounds. During the game, the rule does not change. At the start, the player receives access to five labeled instances (training data). We ensure the probability of each instance to be labeled positive is 50\% (and accordingly 50\% to be labeled negative) to account for imbalances of positive and negative labeled instances in the data set based on the selected rule. Additionally, the player receives five unlabeled instances (test data) that have to be labeled based on the knowledge derived from the labeled training instances. The probability for each instance to be labeled positive remains 50\% as explained before. We then measure the performance on the test data with the \textit{accuracy} metric, which is defined as the number of correctly labeled instances divided by the total number of labeled instances. 

\[ accuracy = \frac{\# \ correctly \ labeled \ instances}{\# \ labeled \ instances} \]

As labeling is only a binary decision in our work, an accuracy indicator of ``1'' is a 100\% correct labeling, whereas an accuracy indicator of ``0.5'' is equivalent to a random guess where labels are randomly assigned. The accuracy of the labeling of the five instances represents the performance in the first round.

\paragraph{Instance} An instance consists of nine elements and a binary label that indicates if the instance fulfills the rule or not.
\paragraph{Round} In every round, humans and machines get five (additional) labeled instances and five new instances to label.  
\paragraph{Game} A game has either 10 (humans) or 20 (machines) rounds.
\paragraph{Experiment} An experiment is finished when four games with four different rules are played.
\newline

\begin{figure}[]
	\centering
	\includegraphics[width=\textwidth]{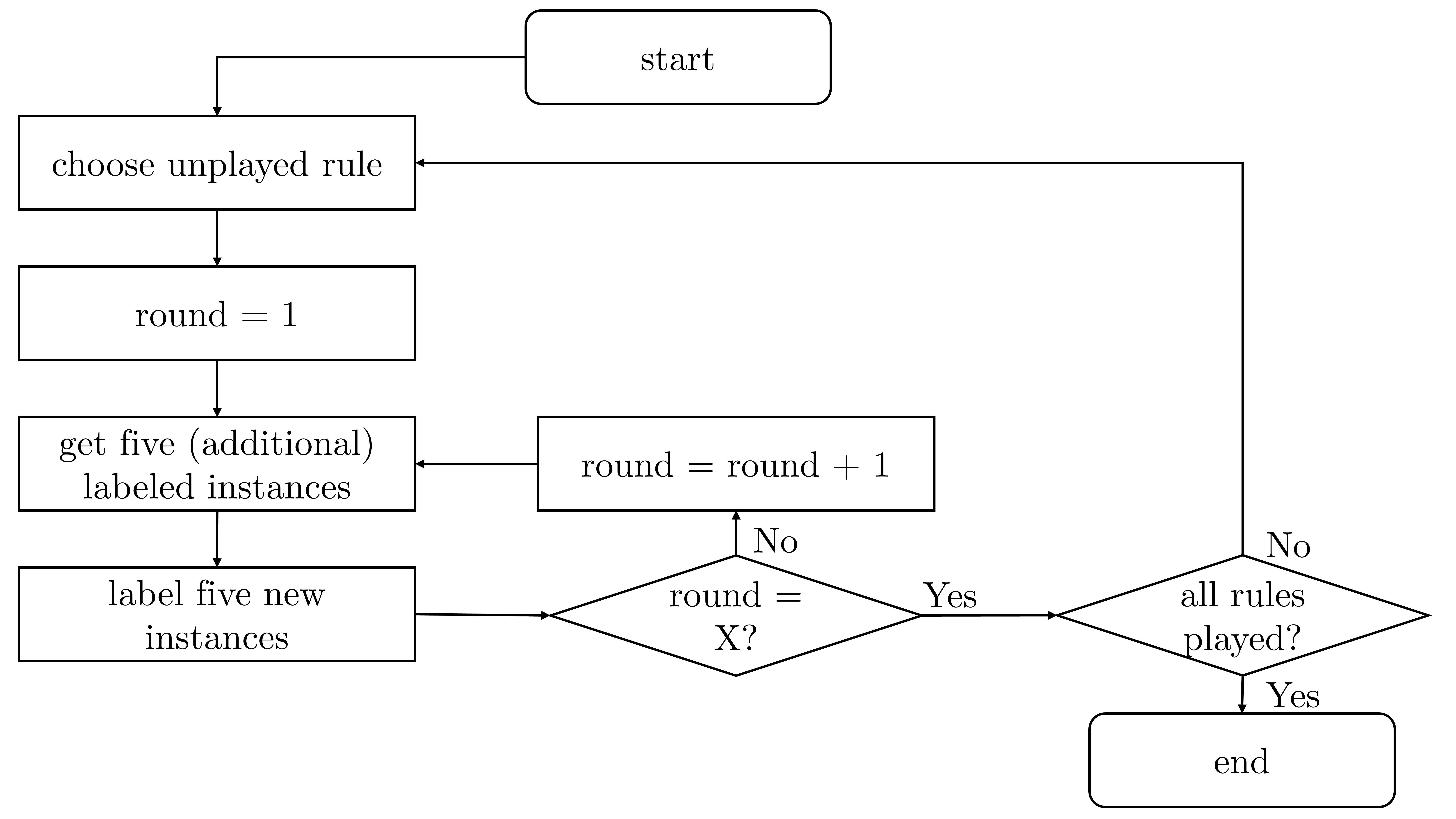}
	\caption{Procedure of the experiment with X = 10 rounds for humans and X = 20 rounds for machines}
	\label{fig:process} 
\end{figure}

In the second round, the previously labeled instances disappear and five new, unlabeled instances are displayed (new test instances). Five additional labeled instances are shown, leading to a total of 10 labeled instances available for training. The labeling of the five new unlabeled instances in the second round determines the performance in Round 2. Evidently, additional rounds follow the same pattern. This is depicted in \Cref{fig:process}. The order of labeled and unlabeled instances is randomized in every game. However, one matrix (instance) will only be part of either the training or testing data, not both. The learning curve is generated based on the performances in each round.

\subsubsection{Experiment with humans}
\label{subsubsec:humanexperiment} 
The experiment with humans is conducted by studying participants in different sessions. They participate in the experiment individually and without any prior knowledge. In advance, they get a standardized introduction about the general aim of the experiment, the layout of the user interface, and some abstract examples. This introduction is available before and during the experiment in printed form, and they can use scrap paper and a pencil to make notes. 
		
\begin{figure}[]
	\centering
	\includegraphics[width=\textwidth]{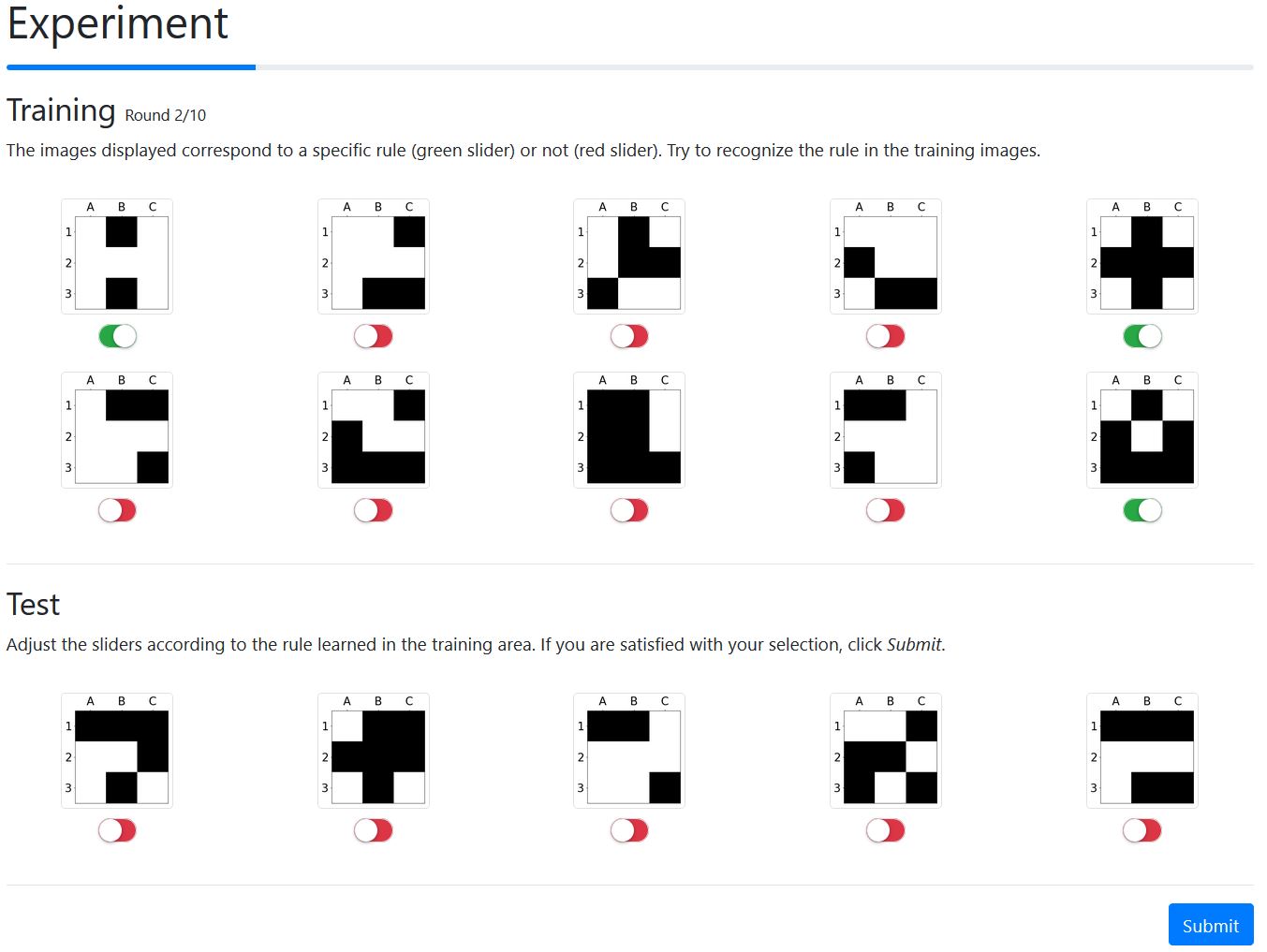}
	\caption{Screenshot of the experiment with humans in Round 2 for the rule \textit{symmetry}. On the top, 10 training instances are displayed and on the bottom part, five unlabeled (test) instances are depicted.}
	\label{fig:screenshot} 
\end{figure}

Every participant has the possibility to play all four rules, leading to four games in total. The total number of rounds per game is limited to 10, which means that the participants will see 50 labeled instances in total and have the opportunity to label 50 instances during one game. After finishing one game, a participant does not receive any feedback about his/her performance. This ensures independent games, as a participant is not influenced regarding the following games. The order of rules is randomized for each participant. \Cref{fig:screenshot} shows an example of GUI of the experiment with humans for the rule \textit{symmetry}.

The experiment is organized and recruited with the software hroot \citep{bock2014hroot}. In total, 44 people participate in two sessions, with 19 people in the first session and 25 in the second one. There are 20 female and 24 male participants, with an average age of 26 years (SD = 9.6). Most experimentees (91\%) are currently enrolled at a German university, majoring in 17 different disciplines, mainly Industrial Engineering and Management (10 students) and Computer Science (seven students).  

Before each session, the experimentees are given instructions on how the experiment works and about their tasks. These instructions are also available on the screen before every game and in printed form during a game. Every person conducts the experiment individually in soundproof cubicles, using a computer. The time limit to complete the experiment with all four rules is set to one hour. The participants are incentivized by an individual payment \citep{kvaloy2015hidden} which is based on their performance relative to that of all other participants in the same session and which ranges from \euro16 (best performance) to \euro7 (worst performance). 
	
\subsubsection{Experiment with machines}
\label{subsubsec:machineexperiment} 

The experiment with machines is conducted by three different algorithms out of all supervised machine learning algorithms, namely a logistic regression (linear), a decision tree (propositional), and a neural network algorithm (non-linear). We do not perform an excessive parameter tuning for the algorithms, because this requires a large number of instances (which is a limitation of our task). This is consistent with the human experiments, because the study participants have no option of playing the game in advance to gain additional knowledge that facilitates completing the task. To increase comparability, every algorithm is applied to every game with the same number of resulting models as the number of humans who played the game. Since we are not limited concerning resources like time, money, or room availability as in the experiment with humans, we can double the number of rounds to 20, which leads to 100 labeled instances (test instances). While our main focus is on the comparison of the first 10 rounds between human and machine, we are curious about how a machine's performance would develop with additional samples. The algorithm is instantiated for one game only and gets terminated after every game so that knowledge from previous games is  not used.

\section{Results}
\label{sec:results} 
After presenting the experiment in the last chapter, we conduct the experiment with humans and the different machine learning algorithms. In this chapter, we evaluate the experiment conducted by humans (\Cref{subsec:resulthuman}) and by the machine learning models (\Cref{subsec:resultmachine}) in detail. In a follow-up step, we compare the human results to those of the machine learning models (\Cref{subsec:compare}).

\subsection{Experiment with humans}
\label{subsec:resulthuman}

The experiment was conducted in two sessions, as described in \Cref{subsubsec:humanexperiment}. Of the participants, 91\% (40 out of 44) finished all four rules within the time limit of one hour. The order of the games is random. We therefore have 42 datasets for the rule \textit{diagonal} and \textit{horizontal} rule and 43 for the rule \textit{numbers} and \textit{symmetry}. 

\begin{table}[htbp]
	\centering
	\begin{tabular}{|l|l|c|}
		\hline
		& P-value  	& Significance   \\ \hline
		Sessions 	& 0.1769   & Not significant \\ \hline
		Rules   	& $1.0\mathrm{e}{-13}$ & $***$            \\ \hline
		Instances 	& $6.6\mathrm{e}{-29}$ & $***$            \\ \hline
		Games   	& 0.0002   & $***$              \\ \hline
	\end{tabular}
	\caption{ANOVA results for session, rules, instances, and games. Significance is depicted with three different levels of statistical significance ($* \ \widehat{=} \ p<0.05$, $** \ \widehat{=} \ p<0.01$, $*** \ \widehat{=} \ p<0.001$), represented with one, two, or three stars.}
	\label{tab:humananova} 
\end{table}

We use analysis of variances (ANOVA) \citep{girden1992anova} to analyze the dataset in a first step (\Cref{tab:humananova}). A one-way ANOVA with the two sessions shows no significance in performance, which indicates that there is no statistical significance between the means of the two sessions. We can therefore analyze all sessions together. To determine the influence of the rules and the number of training instances, we use two-way ANOVAs with replication, since we have a set of paired data where one individual has played several games as well as 10 rounds (= 10 data points) within a game. Since two-way ANOVAs with replication require an equal set of paired data per individual, we exclude the four participants who did not play all four games. Rules and instances show a high statistical significance in performance, as expected in our research question. We account for it later by looking at each rule independently and using learning curves that display the performance for each number of training instances separately without aggregation. 

The performance of the order of games is also statistically significant. This could indicate something like a learning effect between the games, but analyzing this finding is beyond the scope of this article and can be investigated in future research.

\subsection{Experiment with supervised machine learning models}
\label{subsec:resultmachine}
Corresponding to the number of human experimentees who played one rule, we respectively use 42 or 43 machine learning models for each of our three types of machine learning algorithms---regression, decision trees, and neural networks---to play a game for a certain rule. The games are played in the same way that the humans conduct the experiment, seeing five labeled training instances in the first round, and the performance is determined by labeling five instances. As regression algorithm, we choose a logistic regression \citep{scikit-learn} with an L-BFGS solver \citep{liu1989limited}. The used decision tree algorithm is a \textit{DecicionTree Classifier} \citep{Breiman, scikit-learn} and a \textit{Multilayer Perceptron} (MLP) \citep{glorot2010understanding, scikit-learn} with an L-BFGS solver as our neural network algorithm of choice. In the following section, we will use the terms of MLP, decision tree, and logistic regression, knowing that we will always compare the aggregate of 42 or 43 individual performances of these machine learning algorithms.

\subsection{Comparison}
\label{subsec:compare}	
\begin{figure}[htbp]
	\centering
	\includegraphics[width=\textwidth]{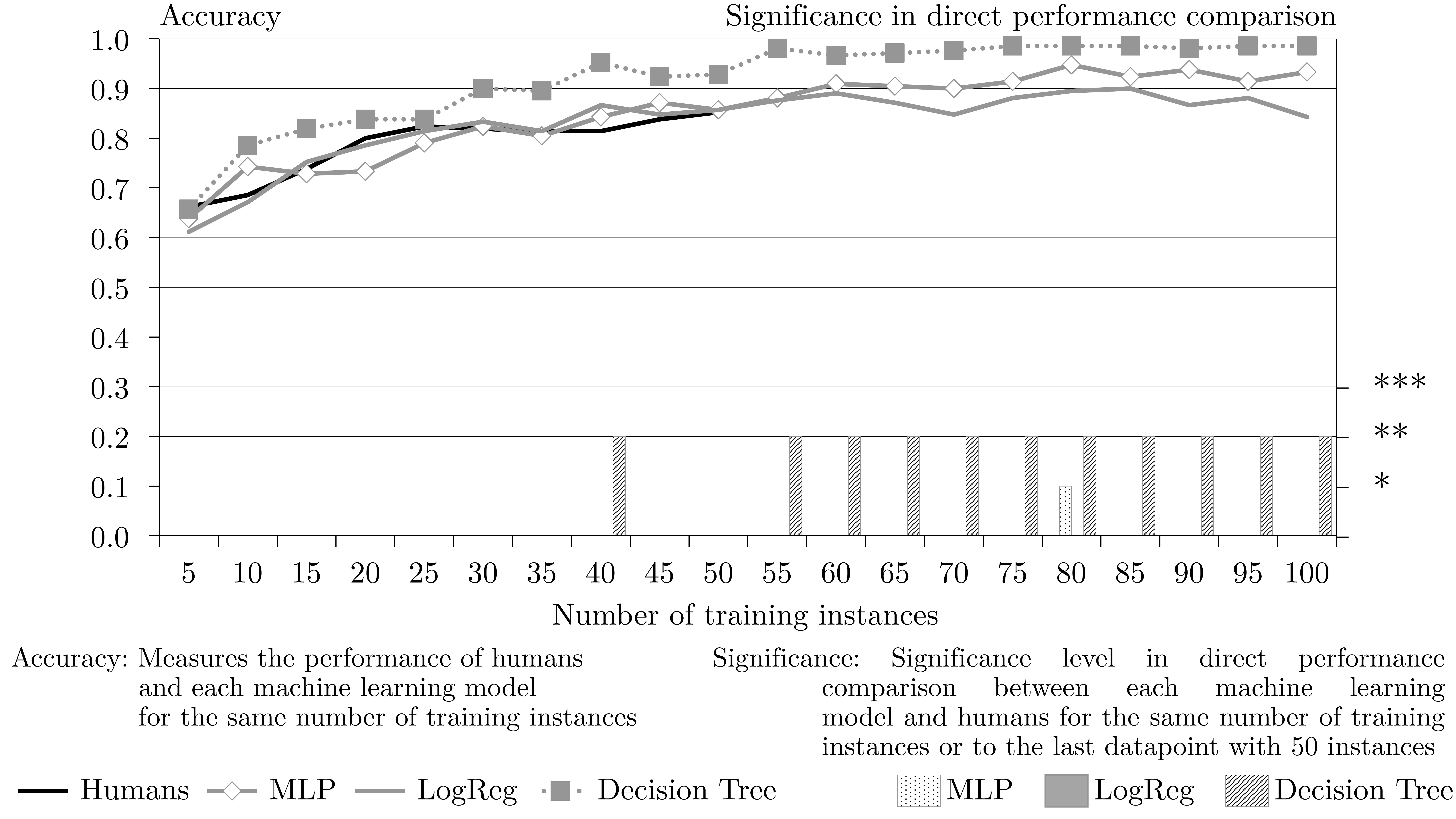}
	\caption{Performance and significances in performance for the rule \textit{diagonal} }
	\label{fig:diagonal} 
\end{figure}

To compare the results of the three machine learning algorithms with the human performance, we analyze each rule through the learning curves our experiment generated. \Cref{fig:diagonal} depicts the results for the rule \textit{diagonal}. The number of training instances is displayed on the x-axis. The left y-axis belongs to the line charts and shows the average accuracy of all experimentees with the given number of training instances. 

The right y-axis belongs to the bar chart. The bar chart shows three different levels of statistical significance ($* \ \widehat{=} \ p<0.05$, $** \ \widehat{=} \ p<0.01$, $*** \ \widehat{=} \ p<0.001$) between the performance of the machine learning models and that of humans. From 55 training instances onward, there is no corresponding human data and the significance refers to the performance difference between machine learning models and humans with 50 training instances. The statistical significance in performance difference is calculated by a two-sided t-test for unequal variance \citep{yuen1974two}. As this results in multiple t-tests on the same dataset, we control the false discovery rate (FDR) by the Benjamini-Hochberg procedure \citep{benjamini1995controlling}.

Regarding the rule \textit{diagonal} (\Cref{fig:diagonal}), the decision tree outperforms all other machine learning models and human participants. In one case, the performance of the decision tree is not significantly better compared to the human within the first 50 training instances. Beginning with 55 training samples, the decision tree performs significantly better ($p<0.01$) than humans in 50 instances. In contrast, the MLP and the logistic regression show similar accuracy compared to the human and do not improve significantly in later rounds. Therefore, those machine learning models do not outperform  humans in 50 training instances significantly. 

\begin{figure}[htbp]
	\centering
	\includegraphics[width=\textwidth]{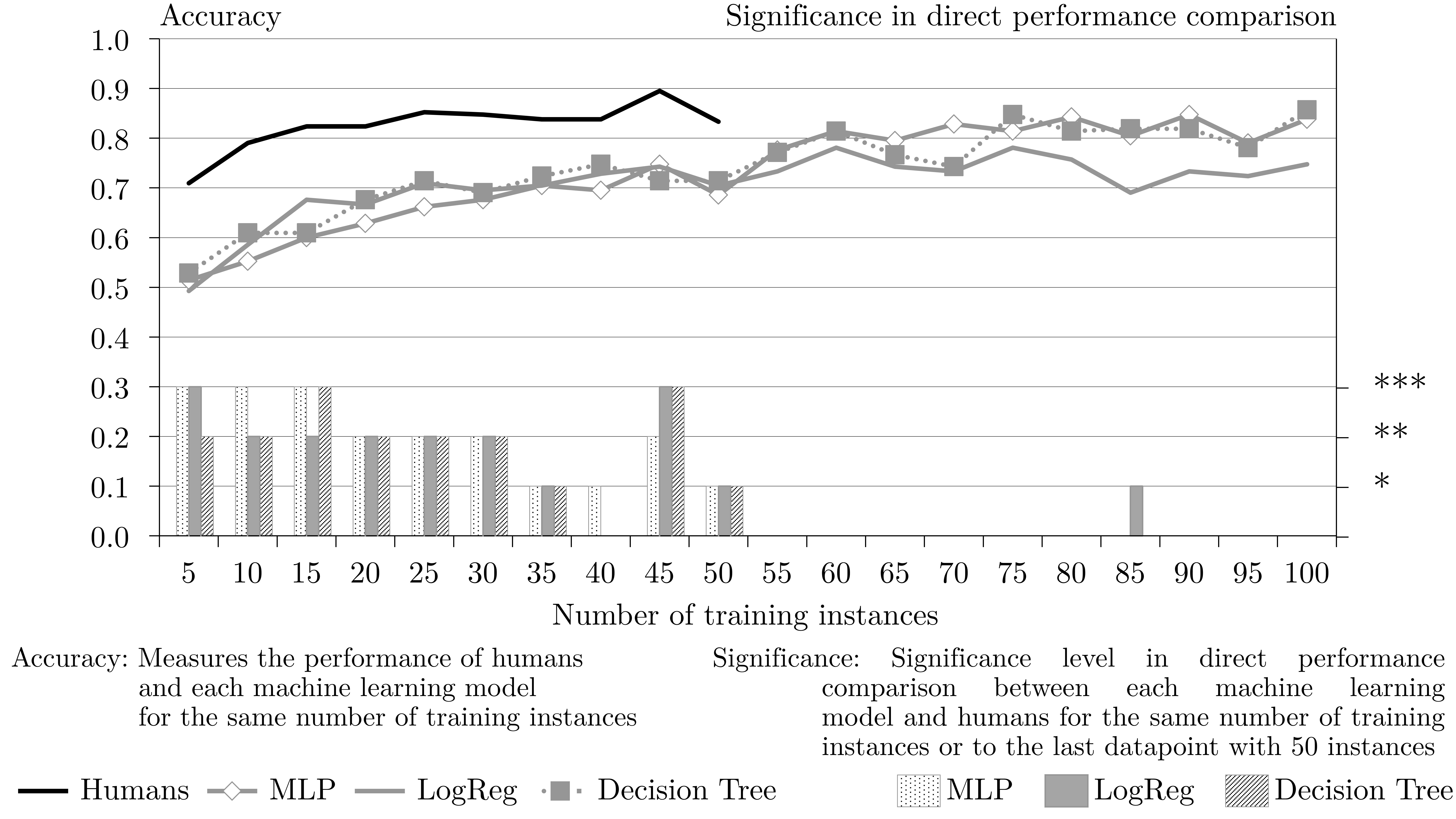}
	\caption{Performance and significances in performance for the rule \textit{horizontal} }
	\label{fig:horizontal} 
\end{figure}

\Cref{fig:horizontal} displays the results for rule \textit{horizontal}. In contrast to the previous chart, humans significantly outperform the machine learning models in the first 50 training instances. However, the statistical significance in performance decreases with more training instances the machine learning models have to learn. Beginning with 55 training instances, the performance of humans with 50 instances and machines with 55 instances does not differ significantly anymore. With the machine learning models, the accuracy is on an equal level and only in the end it seems that the performance of the logistic regression deteriorates a bit.

\begin{figure}[htbp]
	\centering
	\includegraphics[width=\textwidth]{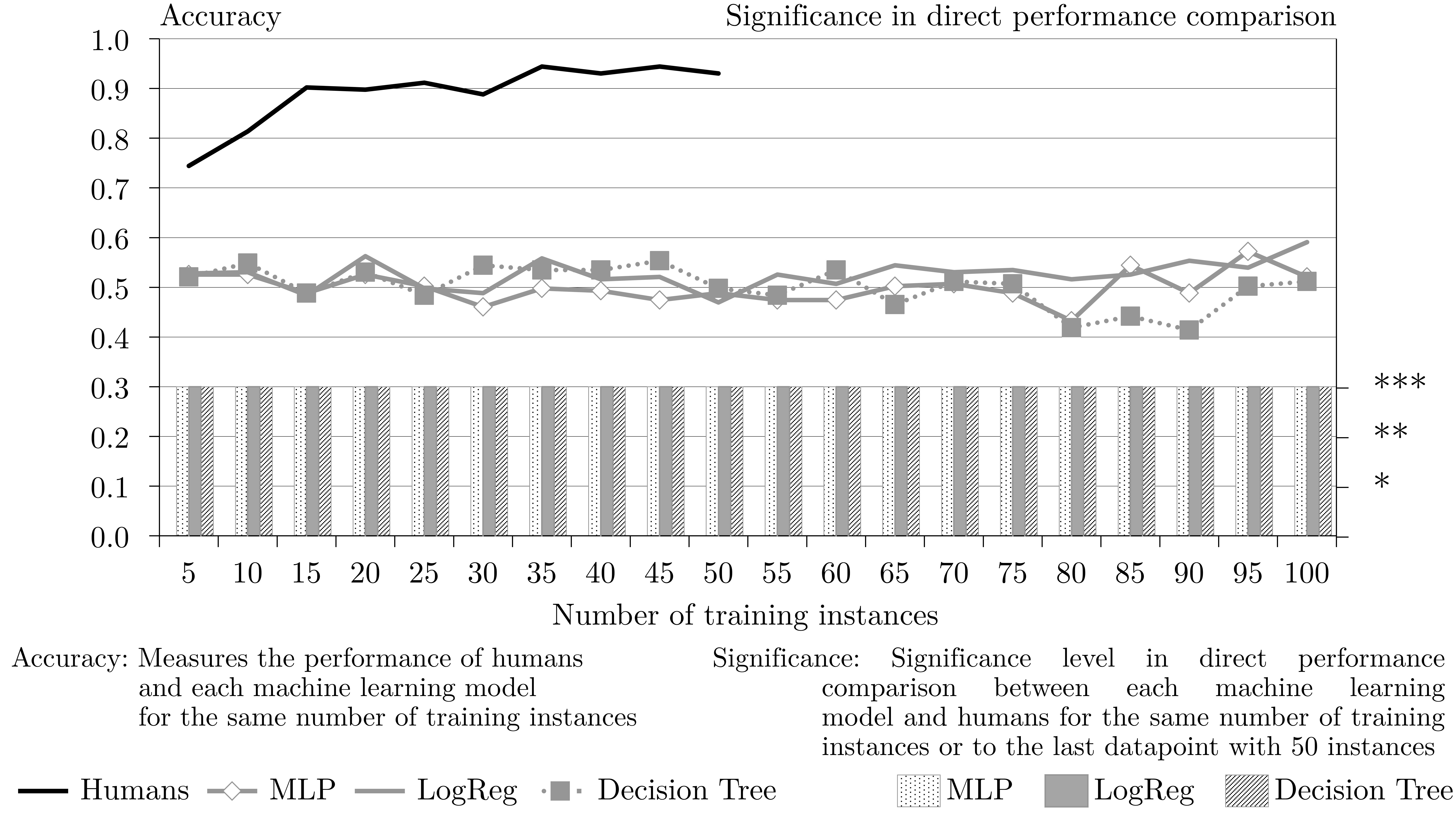}
	\caption{Performance and significances in performance for the rule \textit{numbers} }
	\label{fig:numbers} 
\end{figure}

The rule \textit{numbers} in \Cref{fig:numbers} shows the highest accuracy of human performance across all four rules. Starting with 15 training instances, the performance is always on or above 90\%. The accuracy of the three machine learning models shows no improvement and the accuracy remains at around ``0.5'' for the entire 100 training instances. The performance difference between humans and machine learning models is therefore significant ($p<0.001$) for the experiment across all rounds.

\begin{figure}[htbp]
	\centering
	\includegraphics[width=\textwidth]{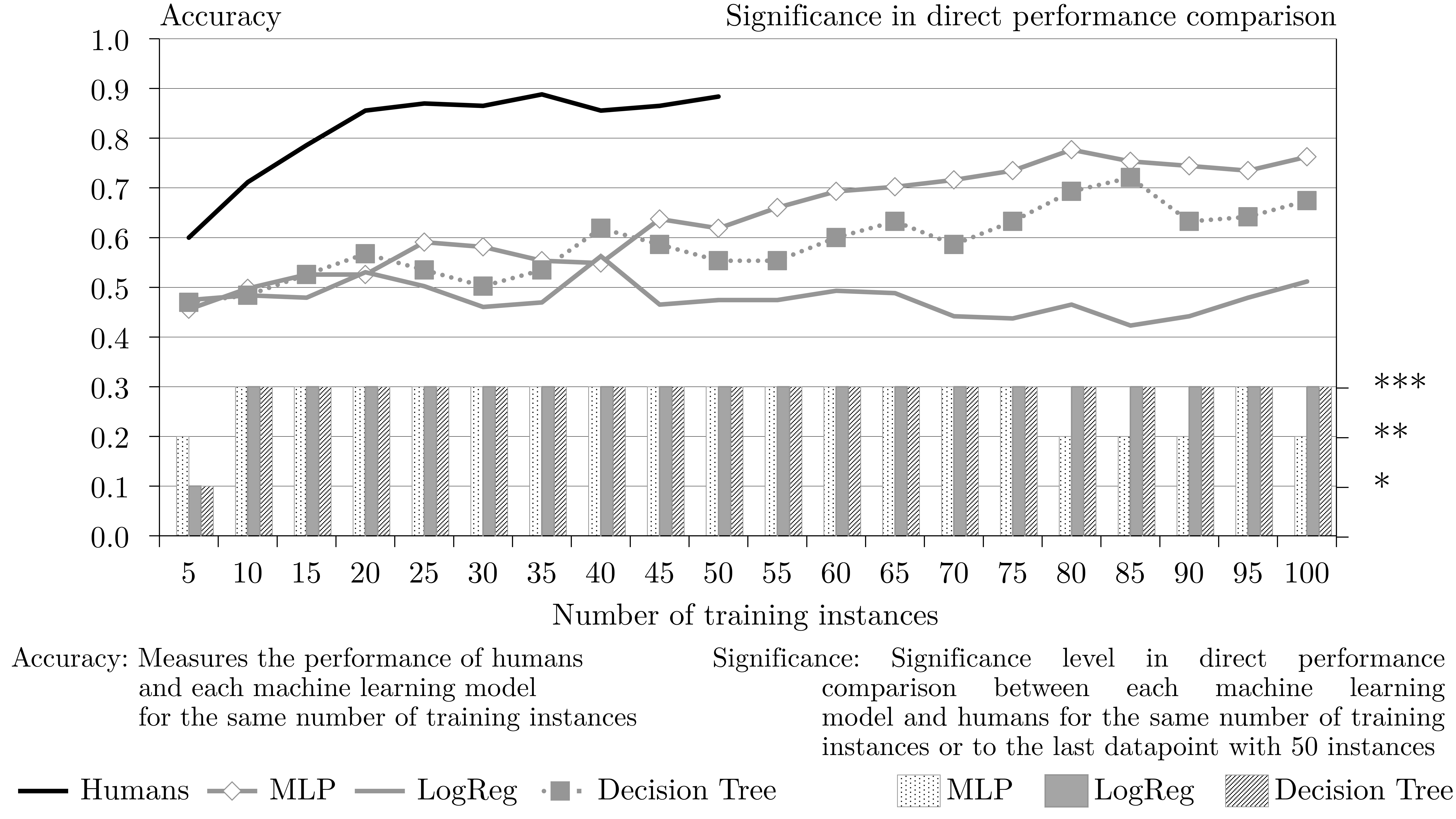}
	\caption{Performance and significances in performance for the rule \textit{symmetry} }
	\label{fig:symmetry} 
\end{figure}

The results for rule \textit{symmetry} are depicted in \Cref{fig:symmetry}. Similar to the human performance in rule \textit{numbers}, humans outperform the machine learning models. With five training instances, the performance is significantly better (MLP: $p<0.01$, decision tree and regression: $p<0.05$)). Afterwards, the human performance improves more than the machine performance and the differences become highly significant ($p<0.001$). However, the human performance reaches its accuracy maximum right below 0.9 after 20 instances and remains on this level, whereas the accuracy of MLP and decision tree slightly improves from round to round. After 50 training instances, the significance level decreases ($p<0.01$).


\section{Discussion}
\label{sec:discussion} 

The results of the experiments provide ground for possible interpretations. We discuss possible explanations of the observed values. This is by no means a full explanation of the shape of every learning curve; however, we give insights on how different theories in computer science and in cognitive psychology can explain some of the outlined results. We start with explanations of the human performance (\Cref{subsec:humandiscussion}) and end with the machine experiment (\Cref{subsec:machinediscussion}).

\subsection{Experiment with humans}
\label{subsec:humandiscussion} 

Given our experiment setup and its results, we employ theories from the area of cognitive psychology, as introduced in \Cref{subsec:humanlearning}, to interpret our results. Other areas of human learning---social cognitive theory and sociocultural theory---are less applicable, as the experiment is performed individually.

The human performance shows two key characteristics across all four rules: High accuracy when labeling the first five instances (no accuracy below 60\%, which outperforms the supervised machine learning models in three of the four rules) and only small performance improvements after learning with 20 or more training instances. 

An explanation for the first observation is grounded in the concept of \textit{one-shot learning} \citep{lee2015neural}. Besides incremental learning, where humans learn step-by-step through trial and error \citep{thorndike1913psychology}, a human is also capable of one-shot learning, which is a technique to learn from a single instance. When a child touches a hot stove plate, he/she will immediately learn not to do it again. This single training instance illustrates one-shot learning. With object recognition, one-shot learning enables humans to recognize objects after one instance by relating the newly seen object to prior knowledge \citep{fei2006one}. Although the used patterns in the experiment are highly abstract, the human can still connect them to known shapes.

The second finding can be explained by cognitive load theory (CLT) \citep{sweller1998cognitive, shaffer2017cognitive, fan2010learning}. CLT describes the learning process as the combination of three loads: the intrinsic cognitive load coming from the difficulty and complexity of the learning subject; the extraneous cognitive load, which originates in preparation of the learning subject; and the germane cognitive load, which describes humans' needed learning capacity to understand the learning subject \citep{paas2003cognitive}. The working memory, where the cognitive process takes place, is essential for the learning process. The working memory itself is limited \citep{ayres2009interdisciplinary, van2005cognitive}, which everyone can experience when playing the board game ``Memory'' and then being unable to memorize every card that has been revealed. Our interpretation of the data suggests that 20 training instances are the limit for humans' working memory, with more training instances only leading to cognitive overload \citep{moreno1999visual} and not to improved performance. An additional explanation can be found in the fatigue effect. The longer the human plays the same rule, the more his performance is effected negatively by fatigue \citep{gonzalez2011cognitive} and counteracts the positive effect of enlarging the data basis by seeing more instances. 

\begin{figure}[htbp]
	\centering
	\includegraphics[width=0.5\textwidth]{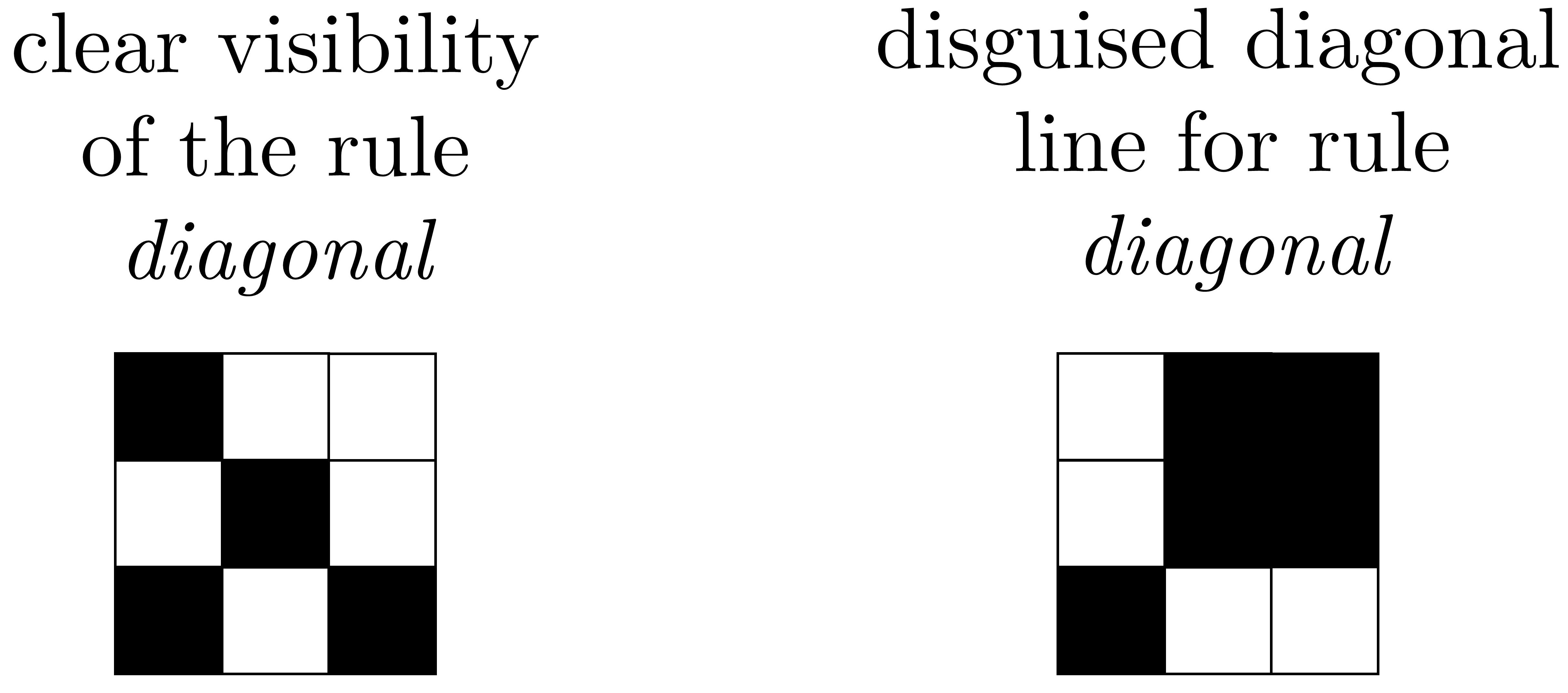}
	\caption{Two instances that fulfill the rule \textit{diagonal}---one with clear visibility of the diagonal line and the other one with two seemingly separated squares of elements disguising the diagonal line}
	\label{fig:optical} 
\end{figure}

Regarding the human performance per rule, the learning curve for the rule \textit{diagonal} is unique. The accuracy in the first round is the second lowest, the maximum performance is the lowest of all four rules, and the machine learning models have similar accuracy numbers or even outperform humans. All these findings indicate that the human performance is particularly bad compared to the other rules. When looking at instances fulfilling the rule \textit{diagonal}, as shown in \Cref{fig:optical}, one can see that the elements (which form the diagonal line) are not joined on the sides but are only linked via their corners. Similar to an optical illusion \citep{coren1978effect}, like the well-known rabbit–duck illustration in which some people see a rabbit and others a duck, the diagonal lines can ``disappear" in some instances while seeing other possible rules. Therefore it becomes harder for humans to see the diagonal line as a potential rule.

Humans show the best performance for the rule \textit{numbers}, which implies that this rule benefits human learning the most. Cognitive fit theory (CFT) \citep{vessey1991cognitive} indicates a link between the task and the chosen type of presentation that leads to superior task performance---the finding of the maximum value in a numerical dataset is completed quicker by humans when plotting the data as a graph compared to a data table. In our experiment, the matrix representation of the data favors counting the number of true elements as well as comparing the counts between training instances.  

Despite the mentioned theories explaining many findings from the results in our experiments with humans, we have to keep the statistical significance in performance between the different games in mind. In future research, we must analyze the results of every game individually and we may find further theories that explain other aspects of human performance.

\subsection{Experiment with machines}
\label{subsec:machinediscussion} 

There are two general findings regarding the machine learning models across all four models: The performance after five training instances is similar or lower compared to the human performance and the machine learning models' performance correlates negatively with the complexity\footnote{In this case, complexity can be understood as the number of constraints / number of basic components needed to describe the pattern; the more basic components, the more complex the problem} of the individual rules.

The first finding relates to the one-shot learning \citep{lee2015neural} we discussed in \Cref{subsec:humandiscussion}. In contrast to humans, all three machine learning models can only do incremental learning. The chosen machine learning models require a certain amount of training instances to perform properly. However, there are special machine learning algorithms designed for one-shot learning, and this is an interesting topic for future work. The Bayesian one-shot algorithm \citep{fei2006one} is an example of a machine learning algorithm that is able to learn via a single instance. 

Regarding the second finding, the complexity of each rule can be defined by the number of basic components. For example, the rule \textit{diagonal} consists of two basic components---either a diagonal line starting in the upper left corner and continuing to the lower right corner, or starting in the lower left corner and ending in the upper right corner. The rule itself leads to the best accuracy numbers for all rules, even outperforming humans. The rule \textit{horizontal} is the combination of three different basic components---either a line in the first, second, or third row. The performance with one additional basic component is slightly lower and gets outperformed for the same number of training instances. The learning curve for the rule \textit{symmetry} shows the third best performance. Because we use an uneven number for the rows and columns of our matrix, the axis on symmetry lies on three elements, which leads to six elements of our matrix being used for the rule. To fulfill the rule \textit{symmetry}, one has to check pairwise whether two elements on opposite sides have the same value, which leads to three pairwise comparisons. By using two different symmetry axis, horizontal and vertical, this rule consists of six basic components. The rule \textit{numbers} can also not simply be broken down into a number of basic components, which may lead to the worst performance of the machine learning models across all four rules. 

Analyzing the machine learning model performances for each rule individually, the decision tree model shows remarkable accuracy for the rule \textit{diagonal}, even outperforming humans. On the one hand, this relates to the comparatively bad performance of humans discussed in \Cref{subsec:humandiscussion}. On the other hand, the rule \textit{diagonal} is unique: The feature $x_5$, referring to the central cell of the matrix, is true---irrespective of the direction of the diagonal line. This circumstance is easy to detect via a decision tree and is a good indication of whether the instance follows the rule or not. 

The rules \textit{numbers} and \textit{symmetry} require the combination of several features and either counting (\textit{numbers}) or comparing features, disregarding their binary status (\textit{symmetry}). A logarithmic regression only looks at each feature individually and fails to detect both rules correctly. In machine learning, the process of feature engineering is utilized frequently---the machine learning model is trained with additional, often human-generated features that are a combination of other (original) features \citep{yu2010feature}. For example, without feature engineering, a decision tree is not able to find the feature count that is necessary for the rule \textit{numbers}. Furthermore, it has problems with detecting differences and ratios, which are essential for the rule \textit{symmetry} \citep{heaton2016empirical}. This may explain the poor performance of the decision tree for the rule \textit{symmetry} and its failure to learn the rule \textit{numbers}. In contrast, a neural network like the used MLP can generate complex features like counting by its layer structure \citep{heaton2016empirical}. The better performance of the MLP compared to the other machine learning models is visible for rule \textit{symmetry}. However, the MLP also fails to learn the rule \textit{numbers} without feature engineering. This may be up to the low number of training instances or an unsuitable default configuration of layers and neurons for the rule \textit{numbers}. 

In accordance with \cite{vapnik1994measuring}, in future work an analysis could be undertaken on the number of examples needed (depending on the class of learner and the class of pattern).

\section{Conclusion}
\label{sec:conclusion} 

This article provides first insights on how learning performance differs between humans and SML models by comparing three different types when there is limited training data. The results of our experiment show a high dependency between performance and the underlying rules of the task. Whereas humans perform relatively similarly across all rules, SML models show big differences between the various patterns. Overall, as expected, humans seem to learn more out of a small number of instances compared to machines. Interestingly, we can observe large differences in the learning curves of our SML models for the different rules we applied in our experiment. In half of the rules we employ, SML models reach the same level or even outperform humans. In the other half, SML models struggle to learn the respective patterns, as those require a deeper understanding that could be gained by a more complex combination of input features, referring to feature engineering. After 20 training instances, humans' performance does not improve anymore in our experiment---arguably due to cognitive overload. Machines learn slower and need more training instances compared to humans.

Our experiment design comes with several limitations: The number of experiment participants could be increased and lead to more, statistically significant results. In addition, we chose three different supervised machine learning algorithms out of hundreds of possible algorithms and parameter combinations. Our selection can only provide a hint of how SML performs in general. The task characteristics have been selected out of a whole set of possibilities. In future, other task combinations need to be used to answer the question on how learning performance differs between humans and machines in general. 

This work shows that further research on the application of supervised machine learning is needed. It is crucial for the application of SML, e.g. as part of autonomous agents, to gain a reliable understanding of tasks and their characteristics that are suitable for automation with SML models. From a business perspective, more research is required on the cost-benefit ratio of replacing human tasks with SML models. This may come with lower task performance, but provides the benefit of automation. We also stress the need for special supervised machine learning algorithms for limited training data, apart from inductive and genetic programming. Continuing the outlined road map of task characteristics and looking at other task characteristic combinations in the future, the single results of each combination will form a more general understanding of the differences between human learning and SML.

\bibliographystyle{elsarticle-harv} 
\bibliography{references}

\end{document}